\documentclass[journal]{IEEEtran}
\usepackage{graphicx}
\usepackage{epstopdf}
\usepackage{multirow}
\usepackage{footmisc}
\usepackage{subfigure}
\usepackage{bbding}
\usepackage{amsmath,lipsum}
\usepackage{color}
\usepackage{url}

\usepackage{fancyhdr}            

\ifCLASSINFOpdf
\else
\fi

\begin{document}
%
\title{DeepSSM: Deep State-Space Model for 3D Human Motion Prediction}
%
%
%

\author{Xiaoli~Liu,~\IEEEmembership{Member,~IEEE,}
        Jianqin~Yin,~\IEEEmembership{Member,~IEEE,}
        Huaping~Liu,~\IEEEmembership{Member,~IEEE,}
        and~Jun~Liu,~\IEEEmembership{Member,~IEEE}
\thanks{Xiaoli Liu and Jianqin Yin are the School of
Artificial Intelligence of Beijing University of Posts and Telecommunications,
No.10 Xitucheng Road, Haidian District, Beijing 100876, China. e-mail: Liuxiaoli134@bupt.edu.cn, jqyin@bupt.edu.cn.}
\thanks{Huaping Liu is with the Department of Computer Science and Technology
of Tsinghua University, Beijing 100084, China.}
\thanks{Jun Liu is with the Department of Mechanical Engineering of City University of
Hong Kong, Hong Kong 999077, China.}
\thanks{Jianqin Yin and Jun Liu are the corresponding authors.}
}

%
%

\markboth{Journal of \LaTeX\ Class Files,~Vol.~14, No.~8, August~2015}%
{Shell \MakeLowercase{\textit{et al.}}: Bare Demo of IEEEtran.cls for IEEE Journals}
%



\maketitle

\thispagestyle{fancy}            
\fancyhead{}                     
\lhead{This work has been submitted to the IEEE for possible publication. Copyright may be transferred without notice, after which this version may no longer be accessible.}                
\cfoot{\quad}                    

\renewcommand{\headrulewidth}{0pt}      
\renewcommand{\footrulewidth}{0pt}

\pagestyle{empty}

\begin{abstract}
Predicting future human motion plays a significant role in human-machine interactions for various real-life applications.
  A unified formulation and multi-order modeling are two critical perspectives for analyzing and representing human motion.
 In contrast to prior works, we improve the multi-order modeling ability of human motion systems for more accurate predictions by building a deep state-space model (DeepSSM). The DeepSSM utilizes the advantages of both the state-space theory and the deep network. Specifically, we formulate the human motion system as the state-space model of a dynamic system and model the motion system by the state-space theory, offering a unified formulation for diverse human motion systems. Moreover, a novel deep network is designed to parameterize this system, which jointly models the state-state transition and state-observation transition processes. In this way, the state of a system is updated by the multi-order information of a time-varying human motion sequence. Multiple future poses are recursively predicted via the state-observation transition. To further improve the model ability of the system, a novel loss, WT-MPJPE (Weighted Temporal Mean Per Joint Position Error), is introduced to optimize the model. The proposed loss encourages the system to achieve more accurate predictions by increasing weights to the early time steps. The experiments on two benchmark datasets (i.e., Human3.6M and 3DPW) confirm that our method achieves state-of-the-art performance with improved accuracy of at least 2.2mm per joint. The code will be available at: \url{https://github.com/lily2lab/DeepSSM.git}.
\end{abstract}

\begin{IEEEkeywords}
Human motion prediction, multi-order modeling, state-space model, deep learning.
\end{IEEEkeywords}

%
\IEEEpeerreviewmaketitle

\section{Introduction}
\begin{figure}[!t]
  \centering
   \includegraphics[width=0.97\columnwidth,height=3.7in,trim = 0mm 0mm 0mm 0mm, clip=true]{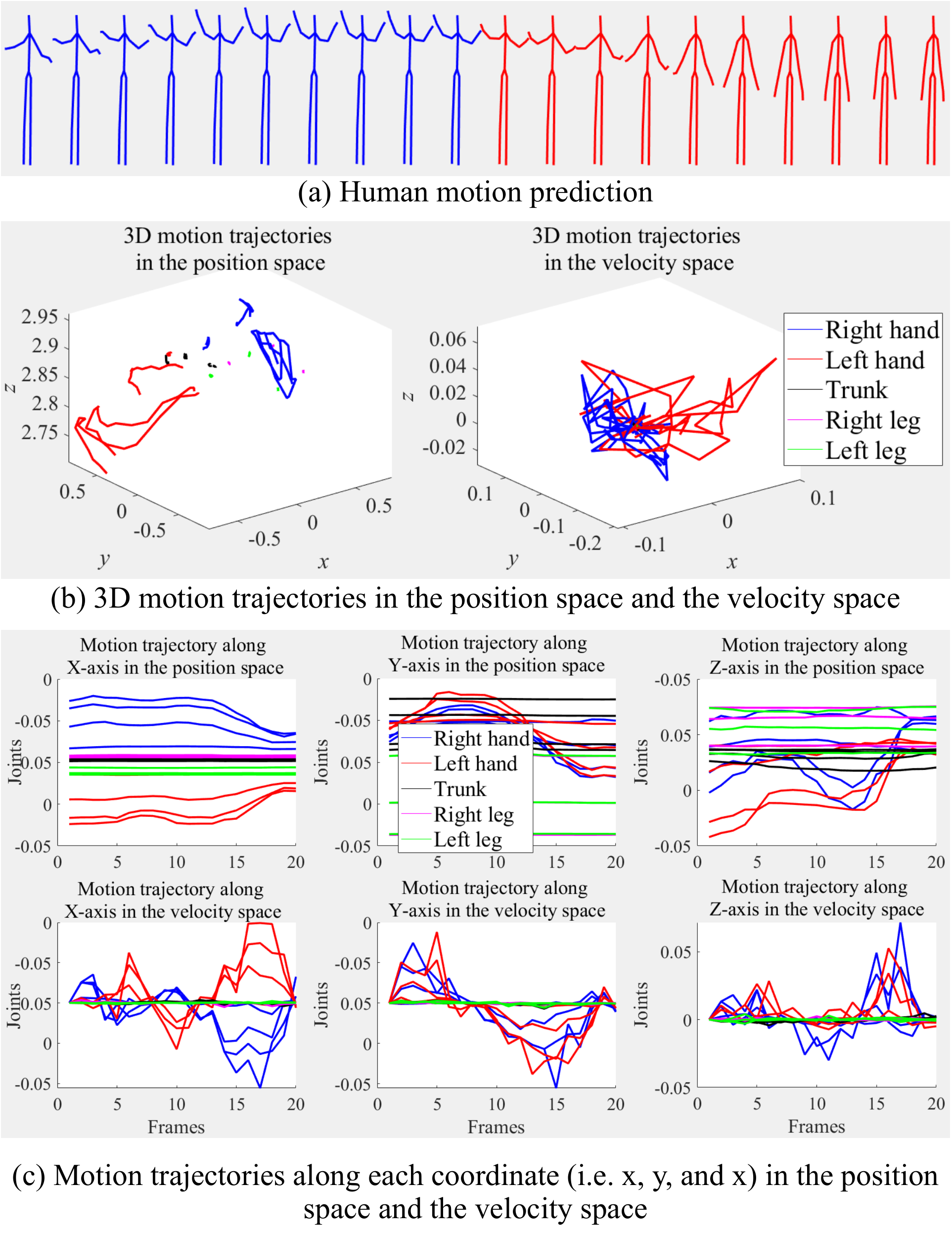}
   \caption{Human motion analysis. ($a$) Human motion prediction, where the blue poses denote the observed poses, and the red poses denote the predictive poses. ($b$) 3D motion trajectories in the position space and the velocity space, where the same color marks joint trajectories of the same limb. ($c$) Joint trajectories along each coordinate (i.e., $x$, $y$ and $z$), where the horizontal axis denotes ``Frames'', the vertical axis denotes ``Coordinate value'' of joints, one curve denotes one joint trajectory along one coordinate, and joint trajectories of the same part are marked with the same color.}
  \label{PP}
\end{figure}

\IEEEPARstart{H}{umans} have a solid ability to make responses according to their surrounding changes over time. \cite{APsurvey,rnmhy}. Similarly, intelligent robots that interact with people must have the ability to predict future dynamics of humans, enabling the robots to respond rapidly to human changes \cite{spl09,pclr19,motiondy17,TIE2019}. In this paper, as shown in Fig. \ref{PP}, we focus on predicting future poses with $3$D joint position data, conditioning on a sequence of observed poses.

Multi-order information, including positions, velocities, etc., carries rich motion dynamics that are useful for predicting human motion. For example, using current positions and future velocities of moving poses, his/her pose at the next time-step can be easily determined \cite{dymod,reviewmotion}. Due to the advantages of multi-order representation, deep networks witness the success of various tasks \cite{sys2013,sys2016,sys2018,xu2019prediction,yu2012predicting}. However, for the human motion prediction, most of the related works focus on modeling the zero-order information (e.g., joint positions) of the pose and neglect the importance of learning higher-order information such as velocities \cite{cstosap,TrajD2019}. Recently, some works have noticed the importance of velocities, and they implicitly model future velocities as the internal state of a human motion system via residual connections in the decoders \cite{hmprnn,li20dyn,19vred}. Moreover, we think the multi-order information of both the input sequence and the output sequence is critical for accurate predictions. Those works in \cite{hmprnn,li20dyn,19vred} ignore the multi-order information of the input sequence, leading to their limited performance. In contrast, we explicitly consider both positions and velocities as the system's inputs and outputs and utilize deep networks to model the multi-order information of the input sequence and output sequence.

Moreover, the human motion system is a typical dynamic system. Formulating it as the state-space model of a dynamic system benefits from two major advantages, i.e.,multi-order modeling and a uniform formalization for a series of diverse human motion systems \cite{SSMlip,deepssm}.
In this paper, we develop a deep state-space model (DeepSSM) that utilizes the merits of both the deep network and the state-space model for predictions. The DeepSSM jointly models two critical processes of the state-space model by the deep network, i.e., the state-state transition and the state-observation transition. Firstly, the positions and velocities of moving poses are incorporated as the observation. The state of the model is initialized to the multi-order information of a sequence of observed poses at both the coordinate level and the joint level by the deep network shown in Fig. \ref{PP}. In this way, according to the state-space theory, the second-order information (i.e., acceleration) is also incorporated into the system.
Secondly, multiple future poses can be predicted recursively via the state-observation transition. Furthermore, the proposed DeepSSM is also general and consequential for providing a theoretical basis for analyzing and interpreting other human motion systems \cite{srnnap,Ntmap,li19symbiotic}.

Most of the related works optimize their models simply using $L2$ \cite{cstosap,hmprnn,rnmhy} or MPJPE (Mean Per Joint Position Error) \cite{TrajD2019,h36m} loss over all poses or all joints. Since the loss value at the early time-steps is smaller than that at the later time-steps, these models implicitly focus on the predictions of later time-steps. This implies that they ignore the relationship between the predictions of early time-steps and later time-steps in a chain model, leading to less accurate predictions. The early predictions easily affect the predictions of later time-steps in a recursive model. Therefore, we propose a novel loss that forces the network to focus on the predictions of early time-steps, encouraging the network to achieve more accurate predictions.

Our main contributions can be summarized as follows.
\begin{itemize}
  \item We formulate the human motion system under the state-space theory of the dynamic system, providing a unified formulation for the existing human motion systems.
  \item Under the state-space theory, we build a new deep state-space model (DeepSSM) parameterized by an encoder-decoder framework (DeepSSM). The proposed DeepSSM utilizes the multi-order modeling ability of the deep network to predict future human motion by jointly learning the state-state transition and state-observation transition of the state-space model end to end.
  \item A novel loss, WT-MPJPE, is designed to optimize the proposed model with increasing weights to the early predictions, guiding the model to more accurate predictions.
\end{itemize}

\section{Related work}
\subsection{Human motion prediction}
Human motion data is a typical type of time-series data, and RNN (Recurrent Neural Network) has shown its strong ability to process time-series data. Therefore, a natural method was proposed based on RNN to predict future human motion \cite{rnmhy,pllsrtd,herna,Tnap}. Fragkiadaki et al. \cite{rnmhy} proposed an encoder-recurrent-decoder (ERD) model by incorporating a nonlinear encoder and decoder before and after LSTM cells. In their model, future poses were recursively predicted via state-state transitions by the inherent recurrent units of LSTM cells. Due to the error accumulation inherently in RNN, these models easily converged to mean poses \cite{Ntmap,GuiAdversarial,vred}. Moreover, human movements are constrained by the physical structure of the human body. Traditional RNN models ignored the spatial correlations among joints of the human body. Recently, other RNN models incorporated some skeletal representations such as Lie algebra representation to model the spatial correlations among joints \cite{Tnap,GuiAdversarial}. Liu et al. \cite{Tnap} proposed a novel model, HMR (Hierarchical Motion Recurrent), to anticipate future motion sequences. The authors modeled the global and local motion contexts using LSTMs hierarchically and captured spatial correlations by representing skeletal frames with the Lie algebra representation.

Another method was proposed based on deep feedforward networks for predicting future human motion \cite{TrajD2019,ButepageDRL}. For example, Butepage et al. \cite{ButepageDRL} learned a generic representation from the input Cartesian skeletal data and predicted future $3$D poses using feedforward neural networks. Mao et al. \cite{TrajD2019} proposed a feedforward model for predicting future 3D poses and also achieved state-of-the-art performance. The authors modeled temporal dependencies of human motion using DCT (Discrete Cosine Transform) and captured the spatial structure information of the human body by representing the joints of the human body as a graph using GCN (Graph Convolutional Network).

Positions and velocities jointly determine the state of the human body at the next time-step to a great extent. However, most of the models mentioned above focused on modeling positions of the human body and ignored the modeling of velocities. Recent works noticed the importance of modeling velocities of human motion \cite{hmprnn,li20dyn,li19symbiotic}. Li et al. \cite{li20dyn,li19symbiotic} and Martinez et al. \cite{hmprnn} introduced a residual connection between the input and output of the decoder to implicitly model velocities as the internal state of their human motion system. Chiu et al. \cite{tprnn} also predicted future poses by modeling the velocities entirely, ignoring the modeling of positions.
Unlike these prior works, we explicitly model positions and velocities or even accelerations of human motion via the state-space theory and deep network by incorporating the positions and velocities as the observation and predicting multi-order information of future poses via a state-observation transition.

\subsection{State-space model for time-series problem}
The state-space model offered a unified formulation for a series of sequential models, e.g., human motion systems \cite{SSMlip,deepvbf}. It can also be applied to analyze the existing sequential models \cite{ButepageDRL,pllsrtd}.
For example, Karl et al. \cite{deepvbf} proposed deep variational Bayes filters under assumptions of the latent state-space model for reliable system identification, and this model also potentially provided system theory for downstream tasks.

Recently, some works are similar to ours by incorporating the deep network and state-space model, utilizing the higher-order modeling ability of the deep network and state-space model \cite{deepssm,DSSMP,deepssm_kdd,deepSMM_ijcai}. Kawamura et al. \cite{deepssm} proposed a deep state-space model by incorporating deep neural networks to address the skeleton-based action recognition, which belonged to a classification problem instead of the regression problem of predicting future human motion. Although those works \cite{DSSMP,deepssm_kdd,deepSMM_ijcai} also belonged to the regression problem, there were some differences. One was that they focused on 1-dimensional regression. The other was that they used recurrent natural network (RNN) to parameterize their state-space model, which was agnostic to the structure. In contrast, the regression of human motion belongs to a multiple-dimensional regression problem, and modeling the spatial structure of the human body is critical.

In this paper, we formulate the problem of human motion by the deep state-space model (DeepSSM), which provides unified formulations for diverse human motion systems. The proposed DeepSSM is parameterized by the convolutional neural network (CNN) instead of the RNN. In this way, we better model spatial correlations among joints of the human body and capture long-term dependencies of human motion.  

\section{Methodology}
\subsection{Skeletal Representation} 
Given an input sequence ${S_{ - {T_1} + 1:0}}=\{ p( - {T_1} + 1),\cdots,p( - 1),p(0)\}$ with a length of $T_1$, where ${p}({t_0}) (t_0 = -T_1+1,\cdots, -1, 0)$ denotes the pose of sequence ${S_{ - {T_1} + 1:0}}$ at the $t_0$-th time-step. The velocities of sequence ${S_{ - {T_1} + 1:0}}$ can be defined as ${V_{ - {T_1} + 1:0}} = \{ v( - {T_1} + 1), \cdots ,v( - 1),v(0)\} $, where ${v}({t_0}) = {p}({t_0}) - {p}({t_0} - 1)$ and $v(-T_1+1)=\{0\}$.
This paper introduces a skeletal representation to capture dynamic features better by representing the input sequence in position and velocity spaces.
In the position space, as shown in Fig. \ref{PP}, since motion trajectories along different coordinates vary greatly, the input sequence ${S_{ - {T_1} + 1:0}}$ is represented by three $2$D tensors, including ${\textbf{\emph{S}}_{1x}}$, ${\textbf{\emph{S}}_{1y}}$ and ${\textbf{\emph {S}}_{1z}}$, to conveniently capture the coordinate-level features, representing trajectories of the sequence ${S_{ - {T_1} + 1:0}}$ along the $x$, $y$, and $z$ axes, respectively.
Similarly, in the velocity space, ${V_{ - {T_1} + 1:0}}$ is represented by three $2$D tensors, including ${\textbf{\emph{V}}_{1x}}$,${\textbf{\emph{V}}_{1y}}$ and ${\textbf{\emph{V}}_{1z}}$, representing the velocity information along the $x$, $y$ and $z$ axes, respectively. In our skeletal representation, the width denotes frames, the height denotes joints, and the order of joints is consistent with \cite{pisepp} to conveniently capture the local characteristic of the human body \cite{pisepp,du2015hierarchical}.

\subsection{Problem formulation} \label{sect3.1}
The human motion system is a typical dynamic system, which can be represented by the state-space model of a dynamic system as equations \ref{dysys1} and \ref{dysys2} \cite{dymod,reviewmotion}.
\begin{equation}
I(t+1) = f_1(I(t),t) + a(t)
\label{dysys1}
\end{equation}
\begin{equation}
O(t) = f_2(I(t),t) + b(t)
\label{dysys2}
\end{equation}
where $I(t)$ and $O(t)$ are the state and observation at time $t$, respectively; $a(t)$ and $b(t)$ are the process and measurement noise, respectively; ${f_1}( \cdot )$ and ${f_2}( \cdot )$ denote the system functions.

In this paper, the positions and velocities of human motion are incorporated as the observation. Human movements are constrained by Newton's law. Due to the continuity of human motion, future poses can be predicted in a short window of time, as shown in Fig. \ref{PP}, conditioning on a sequence of the latest observed poses instead of one static pose. For accurate predictions, the multi-order information of a series of historical poses is set to the state of the state-space model, and $a(t)$ and $b(t)$ are initialized to 0, respectively.
The corresponding future sequence is defined as: ${S_{1:{T_2}}} = \{ p(1), p(2), \cdots, p({T_2})\} $ with a length of $T_2$, and its velocities are ${V_{1:{T_2}}} = \{ v(1), v(2), \cdots ,v({T_2})\}$, where ${ p}({t})$ and $ v(t)$ denote the pose and velocity of sequence ${S_{1:{T_2}}}$ at the $t$-th time-step, respectively.
Therefore, the state $I(t)$ and observation $O(t)$ can be defined as equations \ref{It} and \ref{Ot}, respectively.

\begin{equation}
I(t) = \{ {S_{ - {T_1} + 1:t - 1}},{V_{ - {T_1} + 1:t - 1}}\}
\label{It}
\end{equation}
\begin{equation}
O(t) =\{  p(t), v(t) \}
\label{Ot}
\end{equation}
where ${S_{ - {T_1} + 1:t - 1}} = \{ p( - {T_1} + 1), \cdots ,p(0), \cdots ,p(t - 1)\}$ and ${V_{ - {T_1} + 1:t - 1}} = \{ v( - {T_1} + 1), \cdots ,v(0), \cdots ,v(t - 1)\}$ denote the positions and velocities of human motion before the $t$-th time-step, respectively. The ${S_{ - {T_1} + 1:t - 1}}$ and ${V_{ - {T_1} + 1:t - 1}}$ include both observed poses and predictions before time $t$. $O(0)$ is initialized to $ \{ p(0), 0 \}$.

The state-space model can be considered as a two-stage system, including a state-state transition and a state-observation transition. ($1$) State-state transition: this stage is to update the system state through the system function ${f_1}( \cdot )$ by the multi-order information of a time-varying motion sequence that includes both future poses and historical poses, learning by our proposed network automatically.
($2$) State-observation transition: this stage is to calculate the observation by the system function ${f_2}( \cdot )$ from the current state of the system, which can be learned by our decoder automatically. Moreover, the current positions and future velocities of the human body can determine the positions of the human body at the next time step. Therefore, future poses can be calculated by equation \ref{Ot1}.
\begin{equation}
 p(t) =  p(t - 1) +  v(t)
\label{Ot1}
\end{equation}

\subsection{Deep State-Space Model}
\begin{figure*}[!t]  
  \centering
  \includegraphics[width=1.8\columnwidth,height=2.4in,trim = 38mm 20mm 32mm 18mm, clip=true]{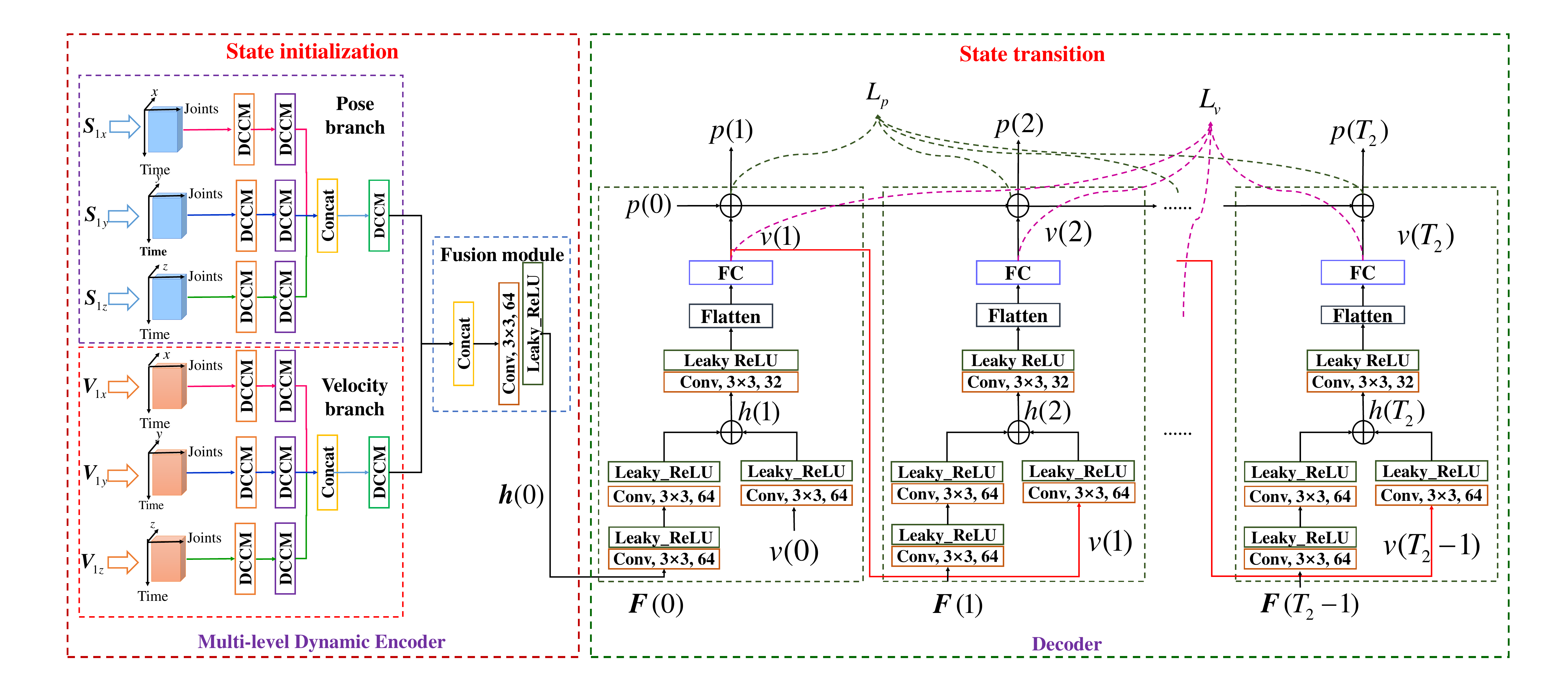}
  \caption{The architecture of DeepSSM, where DCCM denotes our proposed building block shown in Fig. \ref{denseblock}, (Conv, $3 \times 3$, $N$) denotes a $3 \times 3$ convolutional layer with  $N$ output channels. ${\textbf{\emph{F}}}(0) = {\textbf{\emph{h}}}(0)$, ${\textbf{\emph{F}}}(1){\rm{ = }}{\textbf{\emph{h}}}(1)$, ${\textbf{\emph{F}}}(2i + 1){\rm{ = }}{\textbf{\emph{h}}}(2i + 1)$, and ${\textbf{\emph{F}}}(2i){\rm{ = }}{{{H}}_m}([{\textbf{\emph{h}}}(1) + {\textbf{\emph{h}}}(3), \cdots ,{\textbf{\emph{h}}}(2i - 1),{\textbf{\emph{h}}}(2i) + {\textbf{\emph{h}}}(0)])$, where ${{\rm{H}}_m}(\cdot)$ denotes the operation of concatenation across the channel followed by two 3 $\times$ 3
convolutional layers, $i$ is an integer, and $2i{\rm{ + }}1 \le {T_2}$.}
  \label{agvnet}
\end{figure*}

The DeepSSM is parameterized by the new proposed encoder-decoder framework shown in Fig. \ref{agvnet}, including state initialization, state transition, and loss.
The state initialization is parameterized by the encoder, initializing the system's state as the multi-order information of a sequence of observed poses. The decoder automatically learns the state transition to update the system's state and generate future poses.

\begin{figure}[!t]
  \centering
  \includegraphics[width=0.85\columnwidth,height=1.0in,trim = 7mm 7mm 5mm 2mm, clip=true]{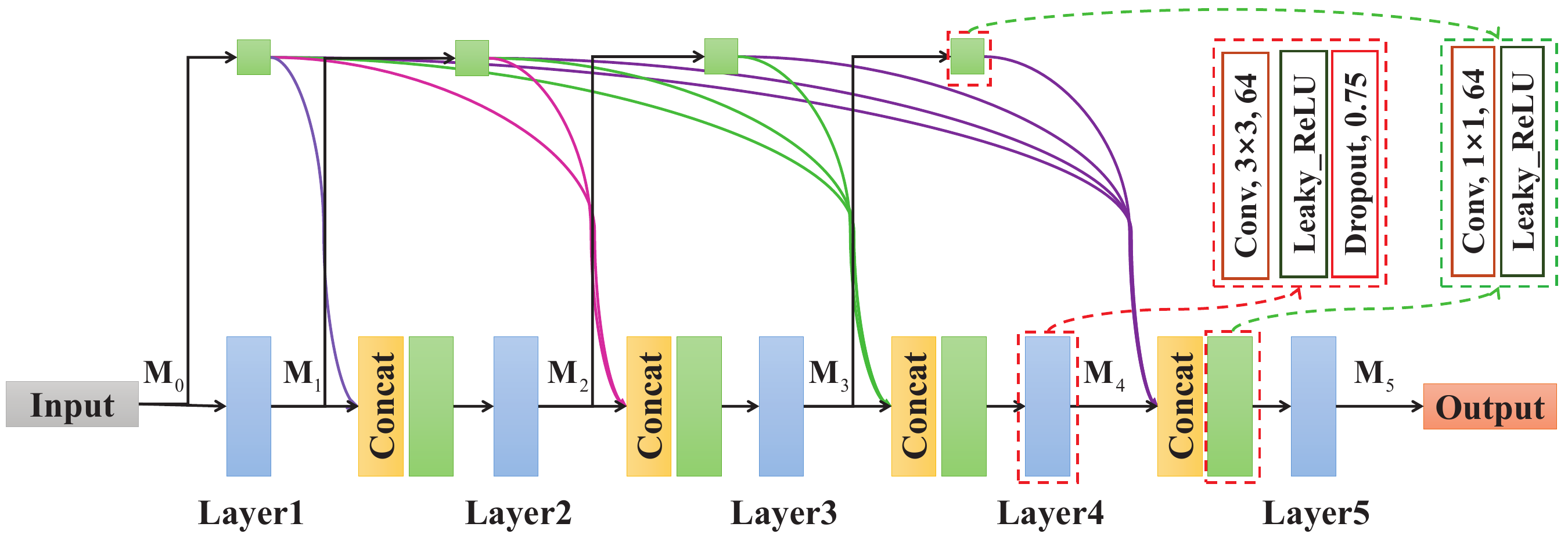}
  \caption{Densely Connected Convolutional Module (DCCM).}
  \label{denseblock}
\end{figure}

{\bf Backbone layer.}
Inspired by \cite{densely}, as shown in Fig. \ref{denseblock}, we propose a new backbone layer, Densely Connected Convolutional Module (DCCM), to maximize the information flow propagation layer by layer, which mainly consists of $5$ convolutional layers. At each convolutional layer, the input receives enhanced features by fusing the concatenated feature maps from all preceding layers using a $1 \times 1$ convolutional layer. Here, the joint-level features can be learned by the $1 \times 1$ convolutions in the residual connections. Therefore, the dense residual connections in DCCM allow the network to gradually obtain the enhanced features of deeper layers by fusing the joint-level features of preceding layers, which can be formulated as equation \ref{eqn7}.

\begin{equation}
{{\bf{M}}_l} = {H_b}([{g_0}({{\bf{M}}_0}),{g_1}({{\mathop{\bf M}\nolimits} _1}), \cdots ,{g_{l - 1}}({{\bf{M}}_{l - 1}}),{{\bf{M}}_{l}}])
\label{eqn7}
\end{equation}
where ${\bf M}_l$ ($l=1,2,3,4$) denotes the output feature map of the $l$-th layer shown in Fig. \ref{denseblock}, ${H_b}( \cdot )$ denotes the fusion layer built with the operation of concatenation across channel followed by a $1 \times 1$ convolution, $g_{*}( \cdot )$ denotes a $1 \times 1$ convolutional layer followed by an activation function (i.e. Leaky ReLU).

{\bf Multi-level dynamic encoder.}
Based on the skeletal representation described above, as shown in Fig. \ref{agvnet}, the multi-level dynamic encoder that includes multi-branch networks is built with DCCMs, which mainly consists of a pose branch, a velocity branch, and a fusion module. The proposed multi-level dynamic encoder models multi-order information of the input sequence at both the coordinate and joint levels in the position and velocity spaces.
($1$) At the pose branch, ${\textbf{\emph S}_{1x}}$, ${\textbf{\emph S}_{1y}}$ and ${\textbf{\emph S}_{1z}}$ are fed into each sub-branch built with $2$ DCCMs, respectively, which enables the network to capture coordinate-level features. Then, one DCCM is applied to obtain joint-level features by fusing coordinate-level features. All sub-branches are shared weights to reduce the model complexity and model correlations among $x$, $y$, and $z$ coordinates.
($2$) At the velocity branch, similar to that of the pose branch, ${\textbf{\emph V}_{1x}}$, ${\textbf{\emph V}_{1y}}$ and ${\textbf{\emph V}_{1z}}$ are fed into each sub-branch respectively to gradually capture the multi-level features of the input sequence in the velocity space.
($3$) The fusion module aims to fuse the features captured by the pose branch and the velocity branch, which is built with the operation of concatenation along the channel followed by a convolutional layer and a Leaky ReLU layer.

{\bf Decoder.}
As shown in Fig. \ref{agvnet}, a recursive feedforward decoder is entirely built with convolutional layers and fully connected (FC) layers. For the decoder, the multi-order information of historical poses (marked as ${{\textbf{\emph F}({*})}}$) has a more complex structure due to multiple poses. In contrast, the current information is relatively simple. Therefore, more operations are applied to the historical features, fewer operations are applied to the current velocity, and an element-wise summation is applied to fuse this information. Finally, another convolutional and FC layers are applied to predict future velocity. At the $t$-th time-step, the decoder receives the multi-order information of historical poses before $(t-1)$ time-step, i.e., ${{\textbf{\emph F}({t-1})}}$, $p(t-1)$, and $v(t-1)$, and outputs the multi-order information at the $t$-th time-step, i.e., $p(t)$ and $v(t)$.
Due to the chain structure of the decoder, the historical features of earlier poses will fade away over time. Therefore,
${{\textbf{\emph F}({t})}}$ is updated by equation \ref{eqnh}, which memorizes long-term dependencies of previous poses before time $t$. Specifically, ${{\textbf{\emph F}({t})}}$ augments the features by sparsely fusing the multi-order information of previous poses via the intermediate outputs ${ \{\textbf{\emph h}(*) \}}$ of decoders.  We have empirically shown no improvement by aggregating the multi-order information of all previous poses except the increasing memory cost.
\begin{equation}
\label{eqnh}
 {\textbf{\emph{F}}}(t) = \left\{
\begin{array}{l}
 {\textbf{\emph{h}}}(t),{\rm{ }}t = 1 \ {\rm{ or }} \ (t \ {\rm{ mod }}\ 2) \ {\rm{ = 1}}\\
{H_m}([{ \textbf{\emph{h}}}(1), {\textbf{\emph{h}}}(3), \cdots , \\ {\textbf{\emph{h}}}(t - 1), {\textbf{\emph{h}}}(t) + \textbf{\emph{h}}(0)]),{\rm{ otherwise}}
\end{array}
\right.
\end{equation}
where ${H_m}( \cdot )$ denotes a memory module built with the operation of concatenation across the channel followed by two $3\times3$ convolutional layers.

{\bf State initialization.}
 As shown in Fig. \ref{agvnet}, the output of the encoder carries multi-order information of the input sequence, and we mark it as $\textbf{\emph h}(0)$. Therefore, the initial state of the proposed DeepSSM is parameterized by the proposed multi-level dynamic encoder.

{\bf State transition.} The state transition, including both the state-state transition and state-observation transition, is updated by equations \ref{eqn9} and \ref{updateob}, respectively.
\begin{equation}
 {I(t+1)}  = {f_1 }(I(t),{t})
\label{eqn9}
\end{equation}
\begin{equation}
O(t) = f_2(I(t),t)
\label{updateob}
\end{equation}
where ${t} = 1,2, \cdots ,T_2$, ${f_1}( \cdot )$ and ${f_2}( \cdot )$ denote two learnable mappings jointly learned by the proposed encoder-decoder framework shown in Fig. \ref{agvnet}.

{\bf Model optimization.}
To model multi-order information of future poses and achieve more accurate predictions, our loss $L$ consists of two parts: ${L_v}$ and ${L_p}$, which can be formulated as: $L{\rm{ = }}{\lambda _1}{L_v}{\rm{ + }}{\lambda _2}{L_p}$, where ${\lambda _1}$ and ${\lambda _2}$ are two hyperparameters to balance the $L_v$ and $L_p$.
($1$) ${L_v}$ guides the network to decode future velocities; ($2$) ${L_p}$ encourages the network to restore future positions.

We propose a Weighted Temporal Mean Per Joint Position Error (WT-MPJPE) loss to avoid error accumulation in a recursive model. The proposed WT-MPJPE pays increasing weights to early predictions, which encourages the network to achieve more accurate predictions at the early time steps. In this way, we can mitigate error accumulation to some extent. Taking $L_v$ as an example, $L_v$ can be defined as equation \ref{eqn10}.
\begin{equation}
{L_v} = \frac{1}{{{N_j}}}\sum\limits_{t = 1}^{{T_2}} {{\alpha _t}\sum\limits_{j = 1}^{{N_j}} {{{\left\| {\widehat J_{jv}^t - J_{jv}^t} \right\|}^2}} }
\label{eqn10}
\end{equation}
where $N_j$ denotes the number of joints, $T_2$ is the number of future poses, ${J_{jv}^t}$ and ${\widehat J_{jv}^t}$ denote the predictive and groundtruth joint in the velocity space, respectively. ${\alpha _t}$ denotes the temporal weight at the $t$-th time-step, and ${\alpha _t} > {\alpha _{t + 1}}$ that forces the network to focus on early predictions. Here, ${\alpha _t}$ is initialized to $2({T_2} - t+ 1)$, and then normalized to $1$ by ${\alpha _t} = \frac{{{\alpha _t}}}{{\sum\nolimits_{k = 1}^{{T_2}} {{\alpha _k}} }}$.

Similar to $L_v$, $L_p$ can be calculated according to equation \ref{eqn10} in the position space.

\section{Experiments}
\subsection{Datasets and Implementation Details}

{\bf Datasets.} (1) {\bf Human3.6M (H3.6M)} \cite{h36m}. H3.6M is the most commonly used dataset for human motion prediction, including 15 actions performed by seven professional actors, such as walking, eating, smoking and discussion. The dataset is collected in control indoor environments with static backgrounds. Each pose is represented by 32 joints of the human body. (2) {\bf 3D Pose in the Wild dataset (3DPW)} \cite{3dpw}. 3DPW is a dataset in the wild with accurate 3D poses performing various activities such as shopping, doing sports. The dataset includes 60 sequences, more than 51k frames, which are collected in outdoor environments with dynamic backgrounds. Each pose is represented by 24 joints of the human body.

{\bf Implementation Details.} All experimental settings and data processing are consistent with the baselines \cite{cstosap,TrajD2019,hmprnn}. The sequences on H3.6M are down-sampled to 25fps, and the sequences on 3DPW are down-sampled to 30fps. For the short-term prediction, $T_1$ and $T_2$ are set to 10 equally; for long-term prediction, $T_1$ and $T_2$ are set to 10 and 25, respectively. Our model is implemented by TensorFlow. MPJPE (Mean Per Joint Position Error) in millimeter (mm) \cite{h36m} is used as our metric to evaluate the performance of our proposed method. All models are trained with Adam optimizer, the learning rate is initialized to 0.0001, and ${\lambda _1}:{\lambda _2}$ is set to 3:1.

\subsection{Comparison with state-of-the-arts}

\begin{table*}[!t] 
\caption{Short and long-term predictions on H3.6M, where the bold marks the best predictions, and the underline marks the second-best predictions.}
\scriptsize
\begin{center}
\begin{tabular}{p{1.1cm}p{0.2cm}p{0.2cm}p{0.2cm}p{0.2cm}p{0.2cm}p{0.4cm}|
p{0.2cm}p{0.2cm}p{0.2cm}p{0.2cm}p{0.2cm}p{0.4cm}|
p{0.2cm}p{0.2cm}p{0.2cm}p{0.2cm}p{0.2cm}p{0.4cm}|
p{0.2cm}p{0.2cm}p{0.2cm}p{0.2cm}p{0.2cm}p{0.3cm}}
\hline
\multirow{2}{*}{Milliseconds}& \multicolumn{6}{c}{Walking} & \multicolumn{6}{c}{Eating}& \multicolumn{6}{c}{Smoking} & \multicolumn{6}{c}{Discussion}\\
\cline{2-25} & 80 &160 & 320 &400 &560 &1000& 80 &160 & 320 &400&560 &1000& 80 &160 & 320 &400&560 &1000& 80 &160 & 320 &400&560 &1000\\
\hline
RGRU\cite{hmprnn} &23.8 &40.4& 62.9& 70.9 &73.8 &86.7& 17.6& 34.7& 71.9& 87.7& 101.3& 119.7& 19.7& 36.6& 61.8& 73.9& 85.0 &118.5&31.7& 61.3& 96.0& 103.5& 120.7& 147.6 \\
CS2S\cite{cstosap} &17.1 &31.2&53.8&61.5& 59.2& 71.3&13.7&25.9&52.5&63.3& 66.5& 85.4&11.1&21.0&33.4&38.3& 42.0& 67.9&18.9&39.3&67.7&75.7 & 84.1& 116.9\\
LTD\cite{TrajD2019}&\underline{8.9} &\underline{15.7}&{\bf 29.2}&{\bf 33.4}& \underline{42.2}& \underline{51.3} & \underline{8.8}& \underline{18.9}& \underline{39.4}& \underline{47.2} &{\bf 56.5}& {\bf 68.6}& \underline{7.8} & \underline{14.9} & \underline{25.3} &{\bf 28.7}& {\bf 32.3}& \underline{60.5} & \underline{9.8} & \underline{22.1} & \underline{39.6} &\underline{44.1}&{\bf 70.4}& {\bf 103.5} \\
Ours&{\bf 7.6} & {\bf 15.6} &\underline{30.2}& \underline{34.9} &{\bf 35.7}&{\bf 48.1}& {\bf 7.8} &{\bf 15.9}&{\bf 33.9}&{\bf 42.5} &\underline{58.5} &\underline{71.4} &{\bf 6.4}&{\bf 13.1} &{\bf 24.2}& \underline{29.6} & \underline{33.0} &{\bf 57.3}&{\bf 8.6} &{\bf 21.3}&{\bf 37.9}&{\bf 43.4}&\underline{70.6} &\underline{109.2}\\
\hline
\hline
\multirow{2}{*}{Milliseconds}& \multicolumn{6}{c}{Directions} & \multicolumn{6}{c}{Greeting}& \multicolumn{6}{c}{Phoning} & \multicolumn{6}{c}{Posing}\\
 \cline{2-25}& 80 &160 & 320 &400&560 &1000 & 80 &160 & 320 &400&560 &1000& 80 &160 & 320 &400&560 &1000& 80 &160 & 320 &400&560 &1000\\
\hline
RGRU\cite{hmprnn} & 36.5 &56.4& 81.5& 97.3&{--} &{--}&37.9& 74.1& 139.0& 158.8 &{--}&{--}&25.6& 44.4& 74.0& 84.2&{--}&{--}& 27.9& 54.7& 131.3& 160.8 &{--}&{--}\\
CS2S\cite{cstosap} & 22.0&37.2 &59.6& 73.4 &{--}&{--}&24.5 &46.2 &90.0& 103.1&{--}&{--}& 17.2& 29.7& 53.4 &61.3&{--}&{--}& 16.1& 35.6& 86.2& 105.6&{--}&{--}\\
LTD\cite{TrajD2019}& \underline{12.6} & \underline{24.4} &\underline{48.2}&\underline{58.4}& \underline{85.8} & \underline{109.3} & \underline{14.5} & \underline{30.5} & \underline{74.2} & \underline{89.0} &{\bf 91.8} &{\bf 87.4} & \underline{11.5} & \underline{20.2} & \underline{37.9} & \underline{43.2} &{\bf 65.0} & {\bf 113.6}&\underline{9.4} & \underline{23.9} & \underline{66.2}&\underline{ 82.9}& {\bf 113.4}& \underline{220.6}  \\
Ours&{\bf 9.7}&{\bf 21.8}&{\bf 47.1}&{\bf 57.5}&{\bf 81.3}&{\bf 104.5}&  {\bf 12.5}&{\bf 27.4}&{\bf 68.0}&{\bf 82.8}&\underline{93.2} &\underline{89.5} & {\bf 10.6}&{\bf 19.2}& {\bf 35.7}&{\bf 42.5} &{\bf 65.0}& \underline{113.7} &{\bf 7.3}&{\bf 21.3}&{\bf 63.9}&{\bf 80.1}&\underline{115.7} &{\bf 210.3}\\
\hline
\hline
\multirow{2}{*}{Milliseconds}& \multicolumn{6}{c}{Purchases} & \multicolumn{6}{c}{Sitting}& \multicolumn{6}{c}{SittingDown} & \multicolumn{6}{c}{TakingPhoto}\\
\cline{2-25} & 80 &160 & 320 &400&560 &1000 & 80 &160 & 320 &400&560 &1000& 80 &160 & 320 &400&560 &1000& 80 &160 & 320 &400&560 &1000\\
\hline
RGRU\cite{hmprnn} & 40.8& 71.8& 104.2& 109.8 &{--}&{--}&34.5& 69.9& 126.3 &141.6& {--}&{--}&28.6& 55.3& 101.6& 118.9 &{--}&{--}&23.6 &47.4& 94.0& 112.7&{--}&{--}\\
CS2S\cite{cstosap} & 29.4& 54.9& 82.2& 93.0 &{--}&{--}&19.8 &42.4& 77.0& 88.4 &{--}&{--}& 17.1& 34.9& 66.3& 77.7&{--}&{--}& 14.0& 27.2& 53.8& 66.2&{--}&{--}\\
LTD\cite{TrajD2019}& \underline{19.6} &{\bf 38.5}&{\bf 64.4}&{\bf 72.2}&\underline{94.3}& \underline{130.4}&\underline{10.7} & \underline{24.6} & \underline{50.6}& \underline{62.0}&{\bf 79.6}&{\bf 114.9}&\underline{11.4} & \underline{27.6}& \underline{56.4} & \underline{67.6} & \underline{82.6} & \underline{140.1} & \underline{6.8} & \underline{15.2} & {\bf 38.2}&{\bf 49.6}&\underline{68.9} &\underline{87.1} \\
Ours&{\bf 17.9}&\underline{39.2} &\underline{64.8} &\underline{75.8} &{\bf 85.9}&{\bf 120.5}& {\bf 9.7}&{\bf 23.3}&{\bf 48.2}&{\bf 61.6}& \underline{82.6} &\underline{116.4} & {\bf 10.3}&{\bf 26.2}&{\bf 51.7}&{\bf 61.3}&{\bf 79.0}&{\bf 131.2}&{\bf  5.2}&{\bf 14.2}&\underline{38.9} &\underline{49.9} &{\bf 68.7}&{\bf 86.8}\\
\hline
\hline
\multirow{2}{*}{Milliseconds}& \multicolumn{6}{c}{Waiting} & \multicolumn{6}{c}{WalkingDog}& \multicolumn{6}{c}{WalkingTogether} & \multicolumn{6}{c}{Average}\\
 \cline{2-25}& 80 &160 & 320 &400&560 &1000& 80 &160 & 320 &400&560 &1000& 80 &160 & 320 &400&560 &1000& 80 &160 & 320 &400&560 &1000\\
\hline
RGRU\cite{hmprnn} & 29.5& 60.5& 119.9& 140.6&{--}&{--}& 60.5& 101.9& 160.8& 188.3&{--}&{--}& 23.5& 45.0& 71.3& 82.8&{--}&{--}& 30.8& 57.0& 99.8& 115.5&{--}&{--}\\
CS2S\cite{cstosap} &17.9& 36.5& 74.9& 90.7&{--}&{--}& 40.6& 74.7& 116.6& 138.7&{--}&{--}& 15.0& 29.9& 54.3& 65.8&{--}&{--}& 19.6& 37.8& 68.1& 80.2&{--}&{--}\\
LTD\cite{TrajD2019}& \underline{9.5} & \underline{22.0} & \underline{57.5} & \underline{73.9} & \underline{100.9} & \underline{167.6} & \underline{32.2} & \underline{58.0} & \underline{102.2} & \underline{122.7} &{\bf 136.6}&{\bf 174.3} & \underline{8.9} & \underline{18.4}& \underline{35.3}& \underline{44.3}& {\bf 57.0}&\underline{85.0}& \underline{12.1} & \underline{25.0} & \underline{51.0} & \underline{61.3} &\underline{78.5} &\underline{114.3}\\
Ours&{\bf 8.1}&{\bf 20.3}&{\bf 52.3}&{\bf 67.0}&{\bf 90.2}&{\bf 162.7} & {\bf 21.9}&{\bf 48.9}&{\bf 89.8}&{\bf 105.4}&\underline{139.8} &\underline{191.8}&  {\bf 6.8}&{\bf 15.7}&{\bf 31.7}&{\bf 41.1}&\underline{58.0} &{\bf 77.7}&{\bf 10.0}&{\bf 22.9}&{\bf 47.9}&{\bf 58.4}&{\bf 77.1}&{\bf 112.7}\\
\hline
\end{tabular}
\end{center}
\label{results_h36m}
\end{table*}
{\bf Baselines}. (1) RGRU \cite{hmprnn} is built entirely based on GRUs, and uses residual connections to predict future velocities implicitly. (2) CS2S \cite{cstosap} is a CNN-based feedforward model and predicts multiple poses recursively. (3) LTD \cite{TrajD2019} is a state-of-the-art method for 3D human motion prediction, which is built with DCT and GCN.

{\bf Results on H3.6M.} We first evaluate our method on the challenging indoor dataset (i.e., H3.6M), and the results for both short-term and long-term predictions are reported in Table \ref{results_h36m}.
Compared with all baselines, our method achieves the best or the second-best performance for both short-term and long-term predictions, showing the effectiveness of our proposed DeepSSM. Specifically, compared with the RNN baseline \cite{hmprnn}, the errors of our method decrease by at least 9.8mm and up to 110.9mm. On the one hand, the proposed DeepSSM utilizes both the deep network and state-space model on multi-order modeling. On the other hand, we explicitly incorporate both positions and velocities as the observation of our DeepSSM. The above reasons make our model better capture multi-order human motion information for superior performance.
In contrast, the RGRU \cite{hmprnn} implicitly models the velocities as the internal state of the system via residual connections in the decoders. The GRU-based model ignores a part of spatial dependencies among joints of the human body and the multi-order information of the input sequence.

Compared with other feedforward baselines \cite{cstosap,TrajD2019}, on average, our model also achieves the best results by a margin of 22mm and 2.2mm, respectively. Firstly, our method explicitly models the positions and velocities as the observation of the proposed deep state-space model, capturing multi-order information at both the encoder and decoder phases. In contrast, the baselines \cite{cstosap,TrajD2019} easily suffer from a limited ability via implicitly velocity modeling by residual connections between the input and output of their decoders. Such models also ignore the multi-order modeling at the encoder phase.
In this way, our method can better capture human motion dynamics using multi-order modeling. Secondly, our recursive decoder incorporating the WT-MPJPE loss enables our network to achieve more accurate predictions. Specifically, our recursive decoder reuses the predictions at the early time steps. Our loss function forces the network to focus on early predictions, which potentially avoids error accumulations and improves the predictions at the later time-steps to some extent. In contrast, LTD \cite{TrajD2019} predicts future poses in a non-recursive manner, which ignores the reusing of early predictions. Furthermore, LTD \cite{TrajD2019} uses a general MPJPE loss to optimize their models, which ignores the predictive difficulty of different time-steps. As a result, these reasons above are likely to make our model achieve superior performance.

\begin{figure}[!t]  
\begin{center}
\subfigure[Greeting]{\includegraphics[width=0.98\columnwidth,height=0.7in,trim = 60mm 18mm 95mm 86mm, clip=true]{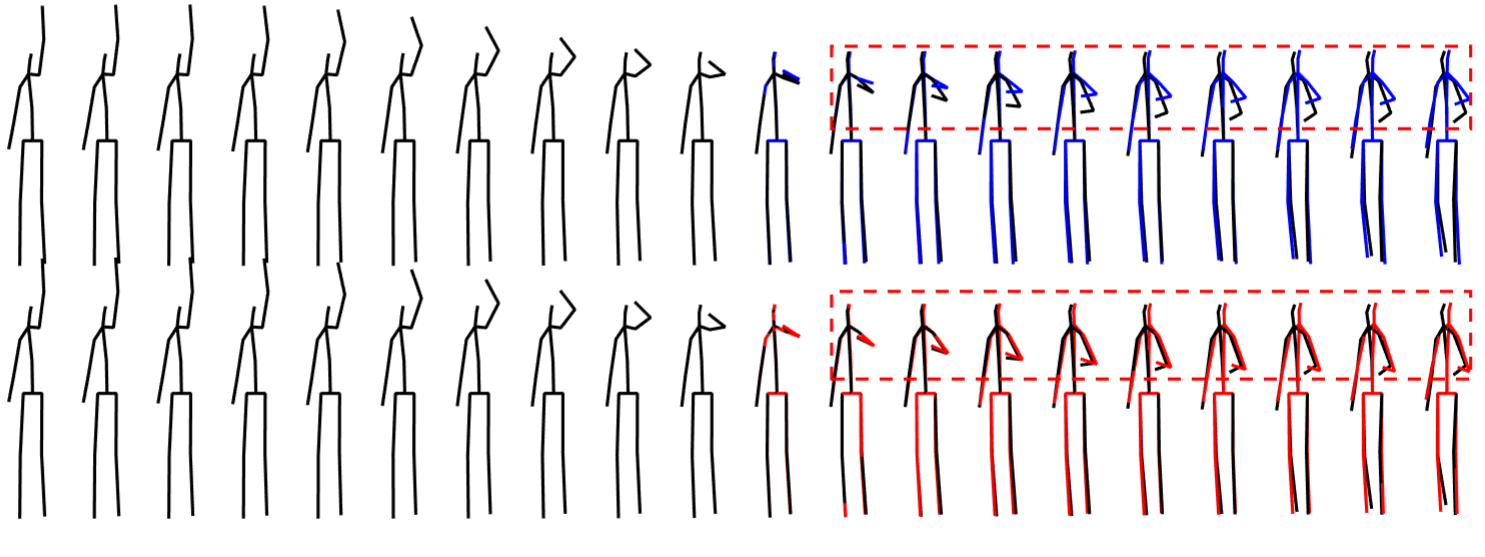}
\label{h36m_vis1}
}
\subfigure[Eating]{\includegraphics[width=0.98\columnwidth,height=0.7in,trim = 60mm 20mm 95mm 125mm, clip=true]{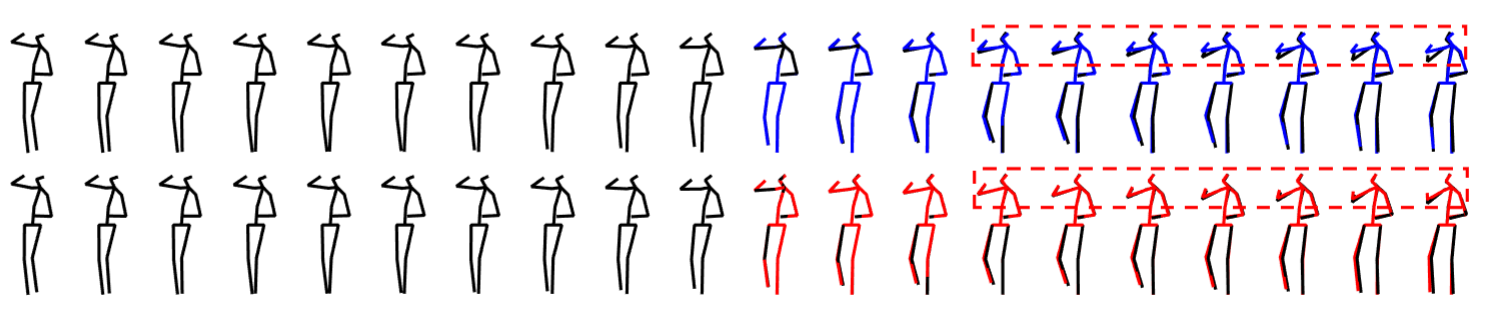}
\label{h36m_vis2}
}
\end{center}
\caption{Qualitative results for short-term predictions on H$3.6$M. The black poses denote the groundtruth, the blue poses denote the results of LTD \cite{TrajD2019}, and the red poses denote the results of our method.}
\label{h36m_vis}
\end{figure}

\begin{figure*}[!t]  
\begin{center}
\subfigure[Sittingdown]{\includegraphics[width=1.96\columnwidth,height=0.8in,trim = 56mm 18mm 80mm 120mm, clip=true]{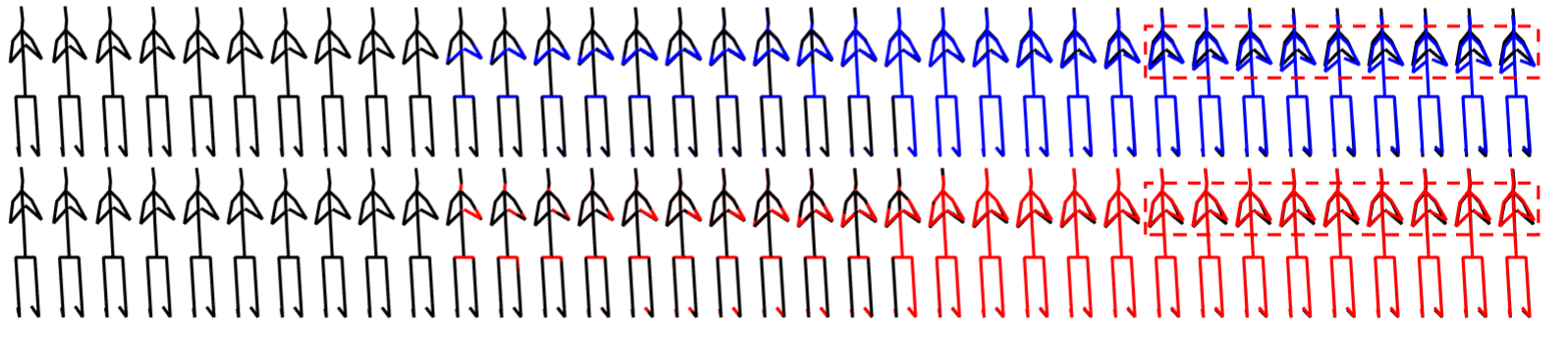}
\label{h36m_vis2_long1}
}
\subfigure[Takingphoto]{\includegraphics[width=1.96\columnwidth,height=0.8in,trim = 52mm 18mm 48mm 29mm, clip=true]{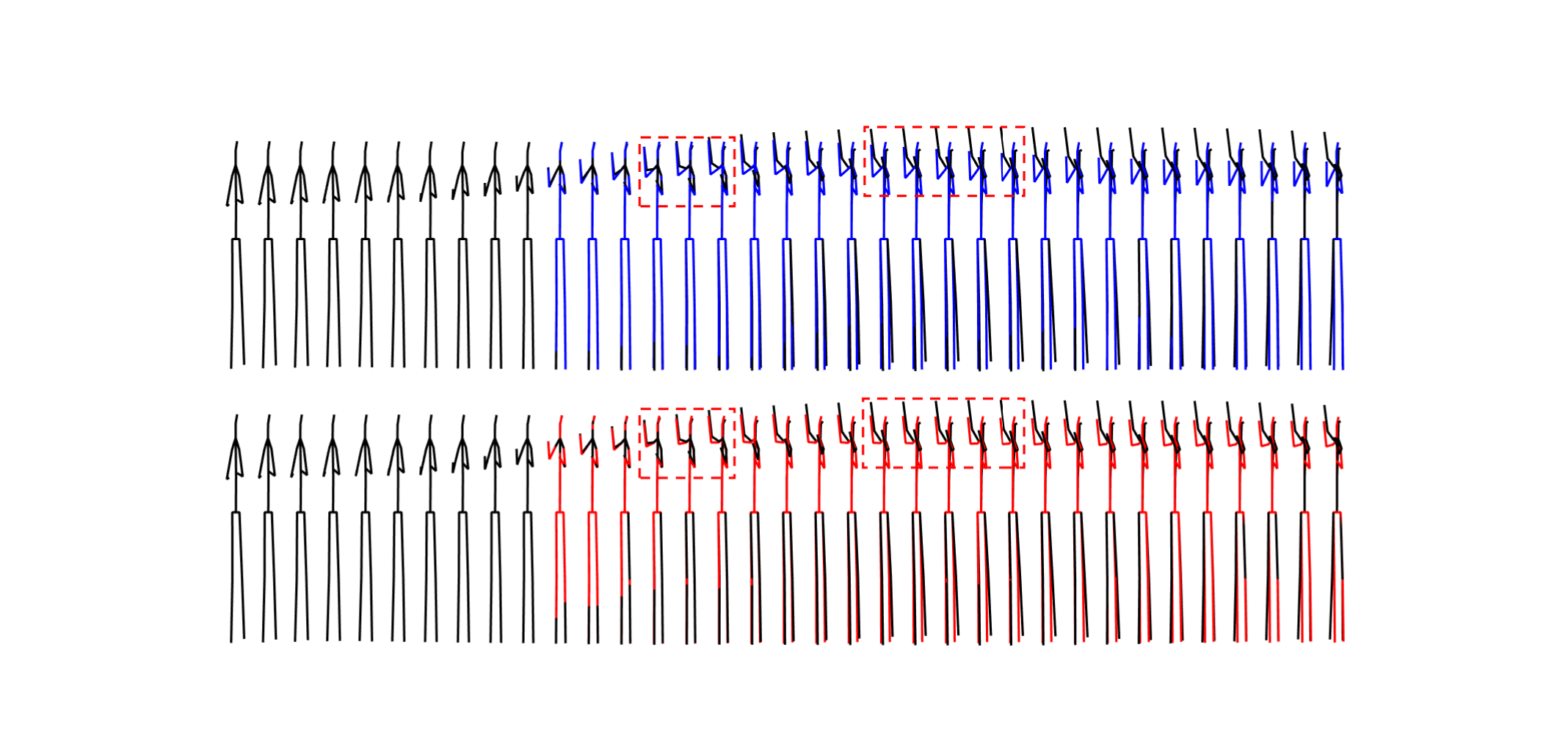}
\label{h36m_vis3_long2}
}
\subfigure[Posing]{\includegraphics[width=1.96\columnwidth,height=0.8in,trim = 58mm 18mm 52mm 125mm, clip=true]{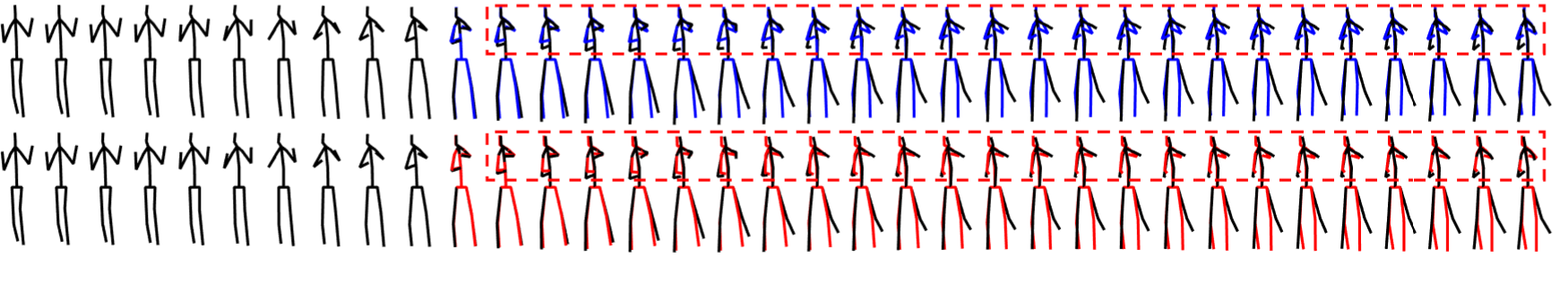}
\label{h36m_vis3_long3}
}
\end{center}
\caption{Qualitative results for long-term predictions on H$3.6$M. The black poses denote the groundtruth, the blue poses denote the results of LTD \cite{TrajD2019}, and the red poses denote the results of our method.}
\label{h36m_vis_long}
\end{figure*}

To further analyze the performance of our proposed DeepSSM, the qualitative results on H3.6M are provided in Fig. \ref{h36m_vis} and Fig. \ref{h36m_vis_long}. Compared with LTD \cite{TrajD2019}, as denoted in the figure, our method also achieves the best visualization performance for both short-term and long-term predictions, demonstrating the effectiveness of our method again. Specifically, for the left hands of Fig. \ref{h36m_vis1} and the right hands of Fig. \ref{h36m_vis2}, the upper limbs of Fig. \ref{h36m_vis2_long1}, and the right hands of Fig. \ref{h36m_vis3_long2} and Fig. \ref{h36m_vis3_long3}, the results of our method are better than that of LTD \cite{TrajD2019}, especially for the fast-moving activities such as the ``Greeting'' in Fig. \ref{h36m_vis1} and the ``Takingphoto'' in Fig. \ref{h36m_vis3_long2}.
In contrast to LTD \cite{TrajD2019}, our method utilizes both merits of deep networks and state-space models. We explicitly model positions and velocities of the human motion, leading to richer multi-order modeling. Moreover, we design a novel temporal loss to encourage the model to achieve accurate predictions. By contrast, LTD \cite{TrajD2019} ignores the modeling of velocities and discards the predictions at the early time-steps, easily suffering from a limited ability of multi-order modeling. Therefore, this may be the possible reason for our model obtaining superior performance.

\begin{table}[!t]
\caption{Short and long-term predictions on 3DPW, where the bold marks the best predictions, and the underline marks the second-best predictions.}
\begin{center}
\begin{tabular}{p{2.0cm}p{0.6cm}p{0.6cm}p{0.6cm}p{0.7cm}p{0.7cm}}
\hline
{Milliseconds} & 200 &400 & 600 &800 & 1000 \\
\hline
RGRU \cite{hmprnn} &113.9 &173.1& 191.9& 201.1& 210.7\\
CS2S \cite{cstosap} & 71.6& 124.9& 155.4& 174.7& 187.5\\
LTD \cite{TrajD2019}&\underline{35.6}& \underline{67.8}&\underline{90.6}& \underline{106.9}& \underline{117.8}\\
 Ours&{\bf 30.3}&{\bf 62.1}&{\bf 88.4}&{\bf 102.1}&{\bf 110.0}\\
\hline
\end{tabular}
\end{center}
\label{3dpwls}
\end{table}

{\bf Results on 3DPW.} We further evaluate our method on the challenging outdoor dataset with dynamic backgrounds (i.e., 3DPW). The results for both short-term and long-term predictions are reported in Table \ref{3dpwls}. In general, our method consistently outperforms the baselines at all time-steps for both short-term and long-term predictions, which further verifies the effectiveness of our proposed DeepSSM. The possible reason for our better performance is our richer multi-order modeling. Moreover, compared with the strong baseline \cite{TrajD2019}, the average performance obtained on this dataset is much higher by a margin of 5.16mm (vs. 2.2mm on H3.6M).

In contrast to the controlled indoor scenes with static backgrounds and minor changes environments on H3.6M, the activities on 3DPW are more challenging because various activities with different performing velocities such as shopping and doing sports and the dynamic backgrounds which include environmental occlusions in the outdoor scenes. On the more challenging dataset, multi-order modeling is more important for accurate predictions. Due to the advantages of multi-order modeling with the proposed DeepSSM, the improvements of our method on this dataset are more significant.

\subsection{Ablation analysis}
In this section, we conduct ablation experiments to thoroughly show the effectiveness of our DeepSSM, including the structure of both the encoder and decoder, multi-order modeling, and WT-MPJPE loss.

\begin{table}[!t]  
\caption{Ablation results of multi-level dynamic encoder on H3.6M, where ``ms'' denotes ``milliseconds'', the bold marks the best predictions, and the underline marks the second-best predictions. ``$xyz$'' denotes modeling the coordinate-level features of poses. ``PV U'' means the pose branch and velocity branch parameters are unshared, but the parameters of the coordinate modeling branches are shared. ``$xyz$ U'' means the coordinate branch parameters are unshared, but each corresponding coordinate branch parameter in the position space and velocity space are shared. ``PV and $xyz$ U'' means that the parameters of all branches (including the pose branch, velocity branch, and their sub-branch for coordinate modeling) are unshared.}
\scriptsize
\begin{center}
\begin{tabular}{p{1.6cm}|p{0.7cm}p{0.7cm}p{0.7cm}p{0.7cm}p{0.7cm}}
\hline
{}&80ms &160ms&320ms&400ms&{Average} \\
\hline
PV U &{\bf 10.0} & \underline{23.0} & 48.7 & 59.6 & 35.3 \\
$xyz$ U & \underline{10.1} & 23.1 & \underline{48.1} & \underline{58.7} & \underline{35.0} \\
PV and $xyz$ U & 10.3 & 23.3 & 48.7 & 59.4&  35.4 \\
\hline
w/o $xyz$& \underline{10.1} & 23.2&49.2&60.1& 35.7\\
\hline
DeepSSM &{\bf 10.0}&{\bf 22.9}&{\bf 47.9}&{\bf 58.4} &{\bf 34.8}\\
\hline
\end{tabular}
\end{center}
\label{abexp2}
\end{table}

{\bf Evaluation of multi-level dynamic encoder.} We first verify the effectiveness of our encoder by exploring the weight sharing scheme among different branches in both the position space and the velocity space. Then we further verify the effectiveness of coordinate modeling. We conduct three experiments for the weight-sharing scheme by respectively removing the weight-sharing scheme in the corresponding branches or sub-branches, denoting by ``PV U'', ``$xyz$ U'' and ``PV and $xyz$ U''.

Without sharing the parameters between the pose branch and the velocity branch (i.e., PV U), the errors increase in most cases, especially at the later time-steps. The results indicate the effectiveness of sharing parameters between the position and velocity spaces. Because the weight sharing scheme potentially explores the correlation among different branches, as shown in Fig. \ref{PP}, the representations of the same human motion sequence in different spaces have certain correlations. Therefore, we can improve the predictive performance. Similarly, without sharing the parameters among $xyz$ coordinates (i.e., $xyz$ U), the performance also declines, showing the effectiveness of modeling the correlations among different coordinates. The performance consistently declines without considering the weight sharing scheme in all branches or sub-branches in the position space or the velocity space (i.e., ``PV and $xyz$ U''). Although the human motion has correlations and constraints among different coordinates, their differences among different axes should not be ignored. As shown in Fig. \ref{PP}, the motion trajectories along different axes are different, and sometimes their movement trends are inconsistent or even opposite. Therefore, without distinguishing these differences (i.e., w/o $xyz$), the model may not capture the high-quality motion dynamics for accurate prediction, showing the importance of considering the modeling of both the coordinate and joint levels in the encoder.

In conclusion, the elegant design in the encoder is reasonable and makes a positive performance for the proposed DeepSSM.

\begin{table}[!t]  
\caption{Ablation results of the decoder on H3.6M, where ``ms'' denotes ``milliseconds'', ``w/o $F(t)$'' denotes not memorizing historical information, ``w/o Sparse $F(t)$'' denotes memorizing the historical information via dense residual connection, i.e., remember all the historical information, and ``LSTM-decoder'' means replacing our feedforward decoder with the LSTM.}
\scriptsize
\begin{center}
\begin{tabular}{p{1.8cm}|p{0.7cm}p{0.7cm}p{0.7cm}p{0.7cm}p{0.7cm}}
\hline
{}&80ms &160ms&320ms&400ms&{Average} \\
\hline
w/o $F(t)$ & {\bf 9.9} & {\bf 22.9} & 48.5 & 59.3 & 35.2 \\
w/o sparse $F(t)$ & \underline{10.0} & \underline{23.0} & \underline{48.1} & \underline{58.5} & \underline{34.9} \\
\hline
LSTM-decoder & 16.3 & 39.9 & 75.2 & 88.8 & 55.1 \\
\hline
DeepSSM &\underline{10.0}&{\bf 22.9}&{\bf 47.9}&{\bf 58.4} &{\bf 34.8}\\
\hline
\end{tabular}
\end{center}
\label{abexp3}
\end{table}

{\bf Evaluation of decoder.} We verify our decoder from those perspectives, and the results are shown in Table \ref{abexp3}. (1) Without memorizing historical information. In this case, we remove the dense connections when updating the $F(t)$. (i.e. w/o F(t)). The state of the next decoder is updated simply by the previous output of decoders, i.e., ${\textbf{\emph{F}}}(t)=h(t)$. As a result, the performance declines, especially for the later time-step. The possible reason is that the information of early poses is easily faded away over time in the recursive model, leading to slightly worse performance. (2) Memorizing all historical information (i.e., w/o sparse $F(t)$). In this case, all historical information is memorized by updating the $F(t)$ with $h(t)$ at all time-steps via dense residual connections. As a result, the performance is comparable to that of DeepSSM. The possible reason is that the historical information between adjacent time-steps is similar. Therefore, it is unnecessary to memorize all historical information. (3) Replacing the proposed feedforward decoder with LSTM (i.e., LSTM-decoder). Similar to that in \cite{hmprnn}, future poses are predicted recursively using LSTM cells. As shown in Table \ref{abexp3}, the errors increase greatly at all time-steps by an average margin of 20.3mm. The results may be caused by the limited ability of spatial correlations among joints of the human body using LSTM-decoder. Instead, our feedforward decoder better captures the spatial information of the human body, and thus we can achieve superior performance.

\begin{table}[!t]  
\caption{Ablation results of multi-order modeling on H3.6M, where ``ms'' denotes ``milliseconds'', ``PI/PF'' denotes the position modeling of the input/future poses, ``VI/VF'' denotes the velocity modeling of the input/future poses, ``P'' denotes the position modeling of both the input and future poses, ``V'' denotes the velocity modeling of both the input and future poses, and ``RVF'' denotes the velocity modeling of the future poses via residual connections between the input and output of the decoder.}
\scriptsize
\begin{center}
\begin{tabular}{p{1.5cm}|p{0.7cm}p{0.7cm}p{0.7cm}p{0.7cm}p{0.7cm}}
\hline
{}&80ms &160ms&320ms&400ms&{Average} \\
\hline
V+PF & {\bf 10.0} & 23.4 & 49.7 & 61.0 & 36.0\\
P+VF & 10.6 & 24.2 & 49.4 & 59.5 &35.9  \\
\hline
V+PI& {\bf 10.0} & \underline{23.2} & \underline{48.4} & 59.4 & 35.3\\
P+VI& 16.1 & 25.4 & 50.3 & 60.9 &38.2\\
P+RVF & 10.7 & 24.2 & 49.2 & \underline{58.9} & 35.8\\
\hline
V &  {\bf 10.0} & 23.6 & 50.3 & 61.8 & 36.4 \\
P & 15.2 & 25.7 & 50.7 & 61.1 &38.2\\
P+VI+RVF & \underline{10.3} & 23.3 & 48.3 & \underline{58.9} & \underline{35.2}\\
\hline
DeepSSM&{\bf 10.0}&{\bf 22.9}&{\bf 47.9}&{\bf 58.4} &{\bf 34.8}\\
\hline
\end{tabular}
\end{center}
\label{abexp1}
\end{table}

{\bf Evaluation of multi-order modeling.} We verify the effectiveness of multi-order modeling of input poses, future poses, and input and future poses. We remove the corresponding branch to ignore the modeling of positions or velocities for the input sequence to achieve this purpose. We remove the corresponding loss to ignore the modeling of future positions and velocities for the future poses. The results are reported in Table \ref{abexp1}.

At the encoding phase, we incorporate both velocities and positions of the input sequence as the observation of the proposed DeepSSM. Whatever the network ignores the modeling of positions or velocities (i.e. ``V+PF'' or ``P+VF''), the errors of DeepSSM decrease in most cases, proving the effectiveness of multi-order modeling.

At the decoding phase, our lose function that includes both $L_p$ and $L_v$ can guide the network to model the multi-order information of future poses (i.e. ``V+PI'', ``P+VI'', and ``P+RVF''). When removing $L_p$, the network ignores modeling positions of future poses and only predicts future velocities. In this way, future poses can be obtained by summarizing positions (i.e., the last observed pose) and velocities of previous predictions, denoting by ``V+PI''. We remove $L_v$ and directly predict the positions of future poses after the FC layers shown in Fig. \ref{agvnet}. In this way, the network completely ignores the velocity modeling of future poses, denoting by ``P+VI''. Furthermore, we simply remove $L_v$ in Fig. \ref{agvnet}. In this case, the network models the positions of future poses. It also implicitly models future velocities using residual connections between the input and the output of decoders, denoting by ``P+RVF''.
Compared with the results of DeepSSM, the errors of ``V+PI'', ``P+VI'' and ``P+RVF'' increase in most cases, showing the effectiveness of multi-order modeling of future poses. Without considering future velocities (i.e. ``P+VI''), the errors increase significantly at all time-steps by a margin of 3.5mm per joint, showing the importance of velocity modeling.

Moreover, we argue that the multi-order information of both input poses and future poses is critical for accurate predictions. For this purpose, we further conduct experiments by simultaneously removing the corresponding parts and the losses of the network (denoted by ``V'' or ``P''). Compared with the results of DeepSSM, the average errors of ``V'' and ``P'' increase by 2.2mm and 3.5mm, respectively, showing the effectiveness of multi-order modeling again. Without considering the velocities in both the encoder and decoder, the performance declines significantly, indicating the importance of velocity learning. The residual connections between the input and output of the decoder can implicitly model the velocities of future poses (i.e. ``P+VI+RVF''). However, the proposed DeepSSM still achieves superior performance by a margin of 0.4mm per joint, further showing the effectiveness of explicitly modeling multi-order information.

In conclusion, we explicitly consider both positions and velocities of human motion as inputs and outputs, enabling the deep network to learn parameters for the proposed DeepSSM better. Therefore, we can further improve the multi-order modeling ability of the system for accurate predictions.

\begin{table}[!t]  
\caption{Ablation results of loss on H3.6M, where ``ms'' denotes ``milliseconds'', and ``WT'' denotes the ``WT-MPJPE'' loss.}
\scriptsize
\begin{center}
\begin{tabular}{p{1.5cm}|p{0.45cm}p{0.45cm}p{0.45cm}p{0.45cm}p{0.65cm}}
\hline
{}&80ms &160ms&320ms&400ms&{Average} \\
\hline
w/o WT & 10.6& 23.7&48.8&59.2&35.6\\
DeepSSM&{\bf 10.0}&{\bf 22.9}&{\bf 47.9}&{\bf 58.4} &{\bf 34.8}\\
\hline
\end{tabular}
\end{center}
\label{abexp4}
\end{table}

{\bf Evaluation of WT-MPJPE loss.} We verify the weighted temporal loss by removing weights for the time-steps (i.e., w/o WT). As shown in Table \ref{abexp4}, the errors of ``w/o WT'' increase at all time-steps, especially at the later time-steps. WT-MPJPE with increasing weights guides the network to predict more accurate results at the early time steps.
Due to the recursive structure, we can potentially mitigate error accumulations and achieve more accurate predictions for the later time steps. As a result, the WT-MPJPE further enhances the overall performance, showing the effectiveness of using different weights for the predictions of different time steps.

\begin{figure*}[!t]  
\begin{center}
\subfigure[Multi-order information of input poses]{\includegraphics[width=0.5\columnwidth,height=2.7in,trim = 2mm 2mm 3mm 1mm, clip=true]{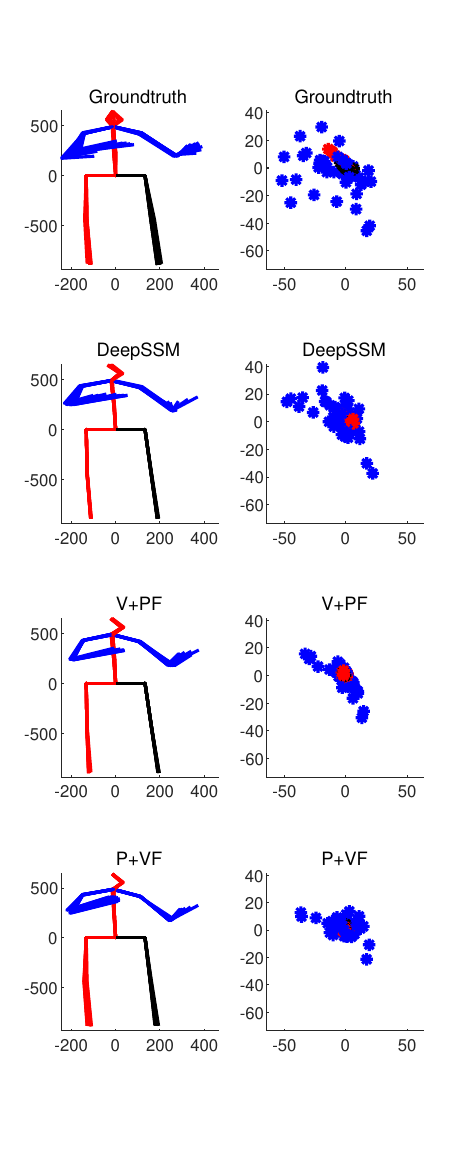}
\label{dirsi}
}
\subfigure[Multi-order information of future poses]{\includegraphics[width=0.5\columnwidth,height=2.7in,trim = 2mm 2mm 3mm 1mm, clip=true]{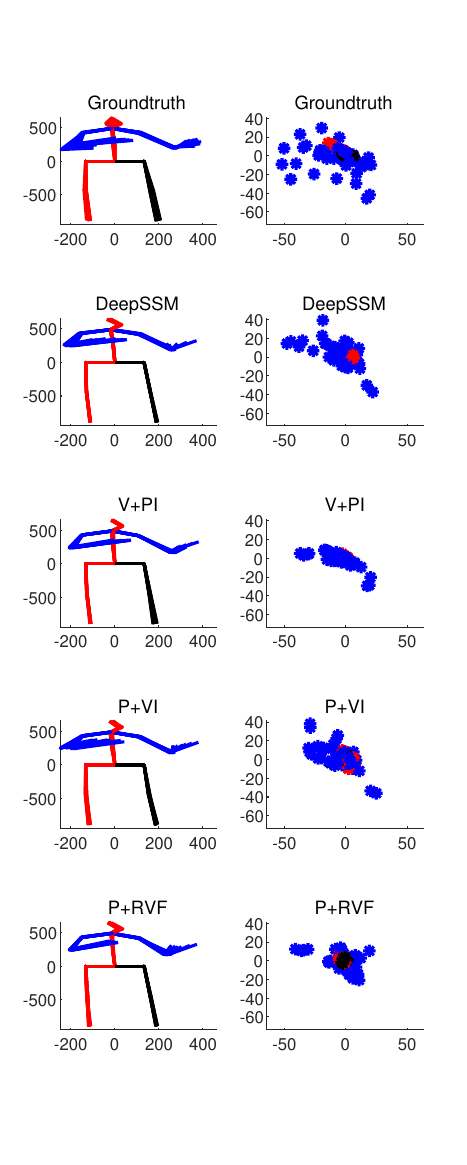}
\label{dirsf}
}
\subfigure[Multi-order information of the whole sequence]{\includegraphics[width=0.5\columnwidth,height=2.7in,trim = 4mm 2mm 3mm 1mm, clip=true]{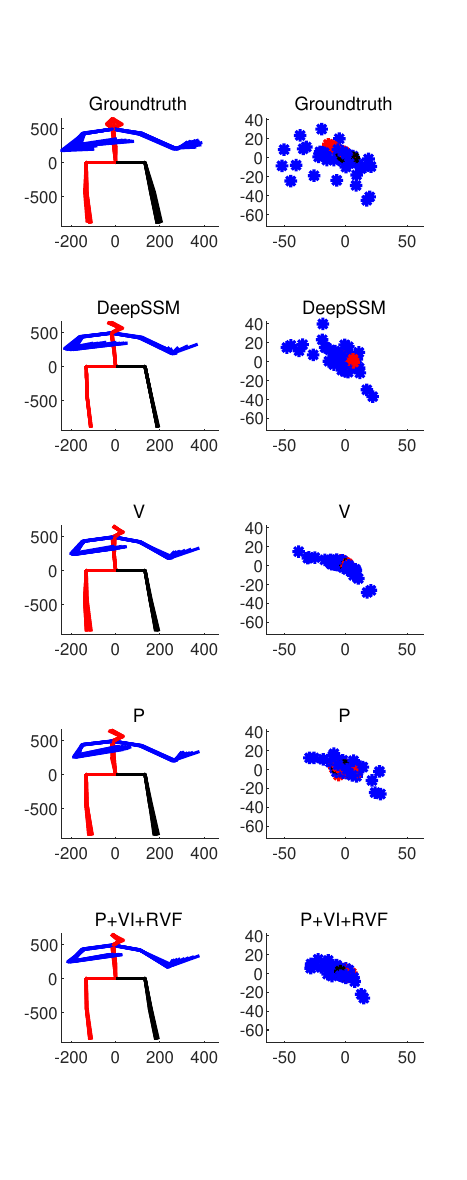}
\label{dirswhole}
}
\end{center}
\caption{Visualizations of the predicted multi-order information which corresponds to Table \ref{abexp1} (Directions). The left ones denote the predicted positions, and the right ones denote the predicted velocities. The blue ones denote the predictions of hands, the red ones denote the predictions of the trunk and right leg, and the black ones denote the predictions of the right leg. It would be noted that when we do not predict future velocities, the velocities of future poses can be obtained by subtraction between the positions of adjacent poses.}
\label{vis_directions}
\end{figure*}

\subsection{Visualizations}
To further show the effectiveness of multi-order modeling, we visualize the multi-order information using the results in Table \ref{abexp1}. The results are shown in Fig. \ref{vis_directions}, where the left ones denote the predicted positions of future poses, the right ones denote the corresponding velocities of future poses. In the figure, different colors denote predictions of different limbs. For the activity of ``Directions'', Fig. \ref{dirsi}, Fig. \ref{dirsf} and Fig. \ref{dirswhole} respectively show the effectiveness of modeling multi-order information of the input sequence, the future sequence and the whole sequence that includes both the input sequence and the future sequence.

For the right hand of the left one in Fig. \ref{dirsi}, ``DeepSSM'' achieves the best predictions, and for the velocities, ``DeepSSM'' obtains the most-similar distributions, showing the effectiveness of modeling multi-order information of the input poses. As shown in Fig. \ref{dirsf}, when we do not consider positions of future poses, the predictions of positions or velocities worse significantly. The phenomenon shows that explicitly modeling future positions greatly helps to predict multi-order information. When we do not consider future velocities, the predictions of hands are getting worse, including the positions and velocities. Although implicitly future velocities via residual connections (i.e. P+RVF) can avoid the poor predictions, the overall performance is still less effective than that of the explicit model one (i.e. DeepSSM) such as the left hand, including positions and velocities. As shown in Fig. \ref{dirswhole}, without considering the positions or velocities of the whole sequence, the performance consistently declines. Compared with ``DeepSSM'', ``P+VI+RVF'' models future velocities implicitly. However, the distributions of predicted hands, including positions and velocities, are dissimilar from ``DeepSSM'' and ``Groundtruth''.

In conclusion, the qualitative results in Fig. \ref{vis_directions} are consistent with the quantitative results in Table \ref{abexp1}, which further verifies the effectiveness of multi-order modeling with our proposed DeepSSM.

\section{Conclusion}
This paper proposes the human motion system as a deep state-space model (DeepSSM). The proposed DeepSSM utilizes the multi-order representation of both the deep network and the state-space model, which provides a unified formulation for various human motion systems. The proposed DeepSSM can also be used to analyze prior models. Furthermore, a novel feedforward-based encoder-decoder is built to parameterize the proposed DeepSSM, jointly achieving state-state transition and state-observation transition. What is more, the proposed WT-MPJPE loss can further guide the recursive model to achieve more accurate predictions. Finally, we evaluate our model on two challenging datasets, achieving state-of-the-art performance. The experiments also show that the coordinate-level features of human motion can further improve the system's performance.

\section*{Acknowledgment}

This work was supported partly by the Fundamental Research Funds for the Central Universities (Grant No. 2020XD-A04-2), partly by the National Natural Science Foundation of China (Grant No. 61673192), and partly supported by BUPT Excellent Ph.D. Students Foundation (CX2021314).

\bibliographystyle{IEEEtran}
\bibliography{DeepSSM}

\ifCLASSOPTIONcaptionsoff
  \newpage
\fi



%

%

\begin{IEEEbiography}[{\includegraphics[width=1in,height=1.25in,clip,keepaspectratio]{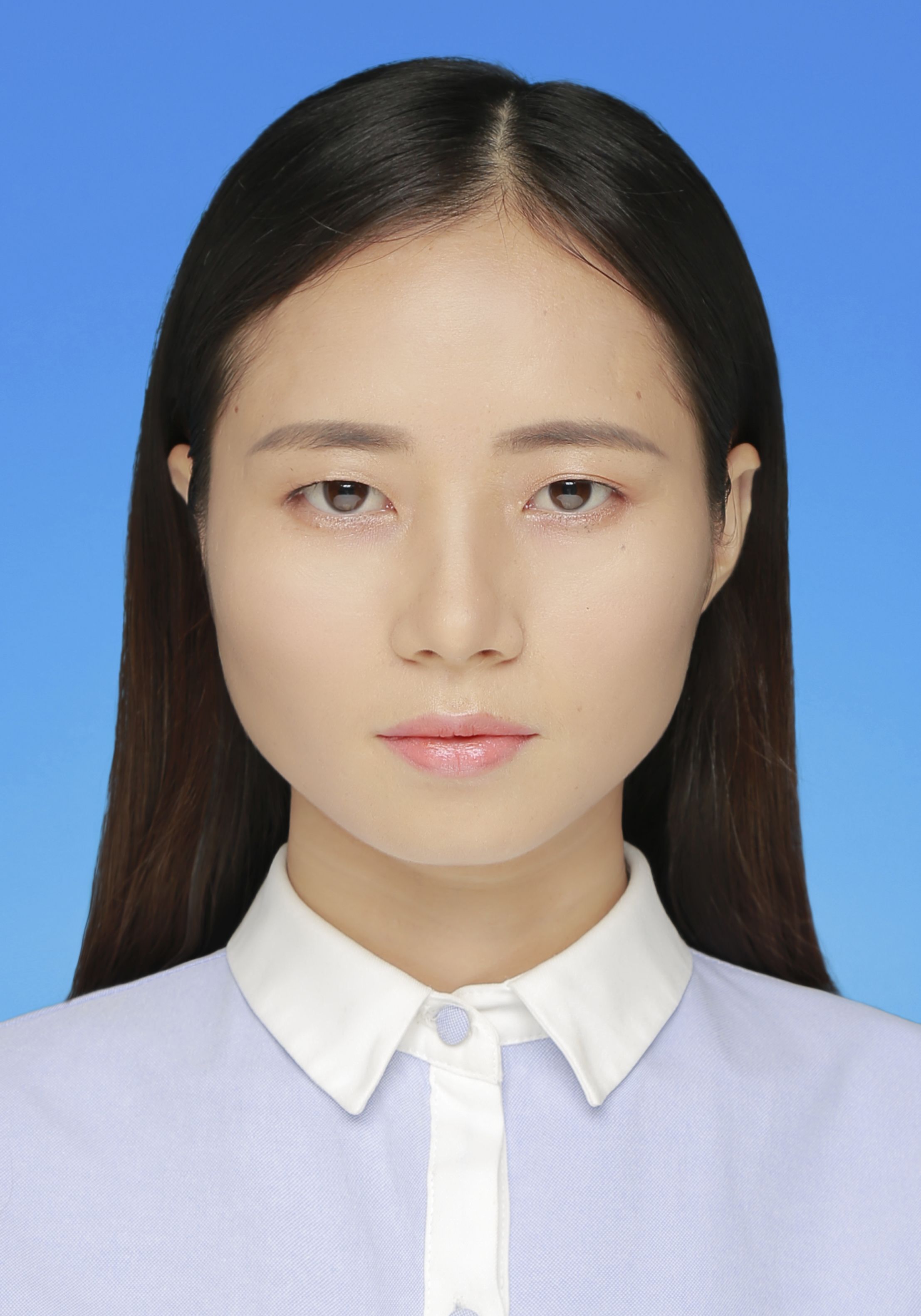}}]{Xiaoli Liu}
Xiaoli Liu received her M.Eng. degree and Bachelor's degree from the University of Jinan, Jinan, China, in 2018 and 2015, respectively. She is currently a Ph.D. candidate with the School of Artificial Intelligence, Beijing University of Posts and Telecommunications, Beijing, China. Her research interests include computer vision, machine learning and image processing, deep learning. Email: Liuxiaoli134@bupt.edu.cn.
\end{IEEEbiography}

\begin{IEEEbiography}[{\includegraphics[width=1in,height=1.25in,clip,keepaspectratio]{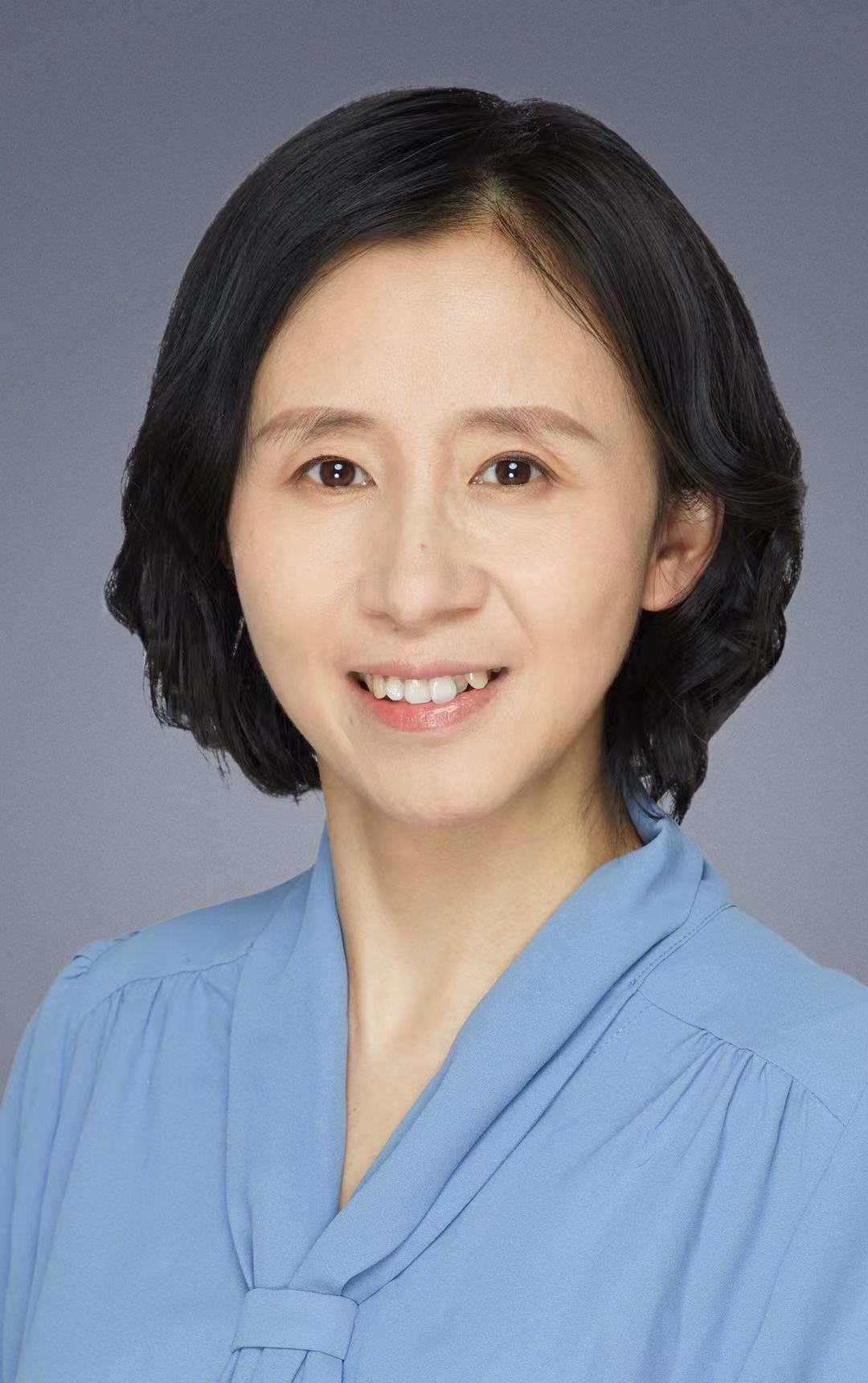}}]{Jianqin Yin}
Jianqin Yin received the Ph.D. degree from Shandong University, Jinan, China, in 2013. \\She currently is a Professor with the School of Artificial Intelligence, Beijing University of Posts and Telecommunications, Beijing, China. Her research interests include service robot, pattern recognition, machine learning and image processing. Email: jqyin@bupt.edu.cn.
\end{IEEEbiography}

\begin{IEEEbiography}[{\includegraphics[width=1in,height=1.25in,clip,keepaspectratio]{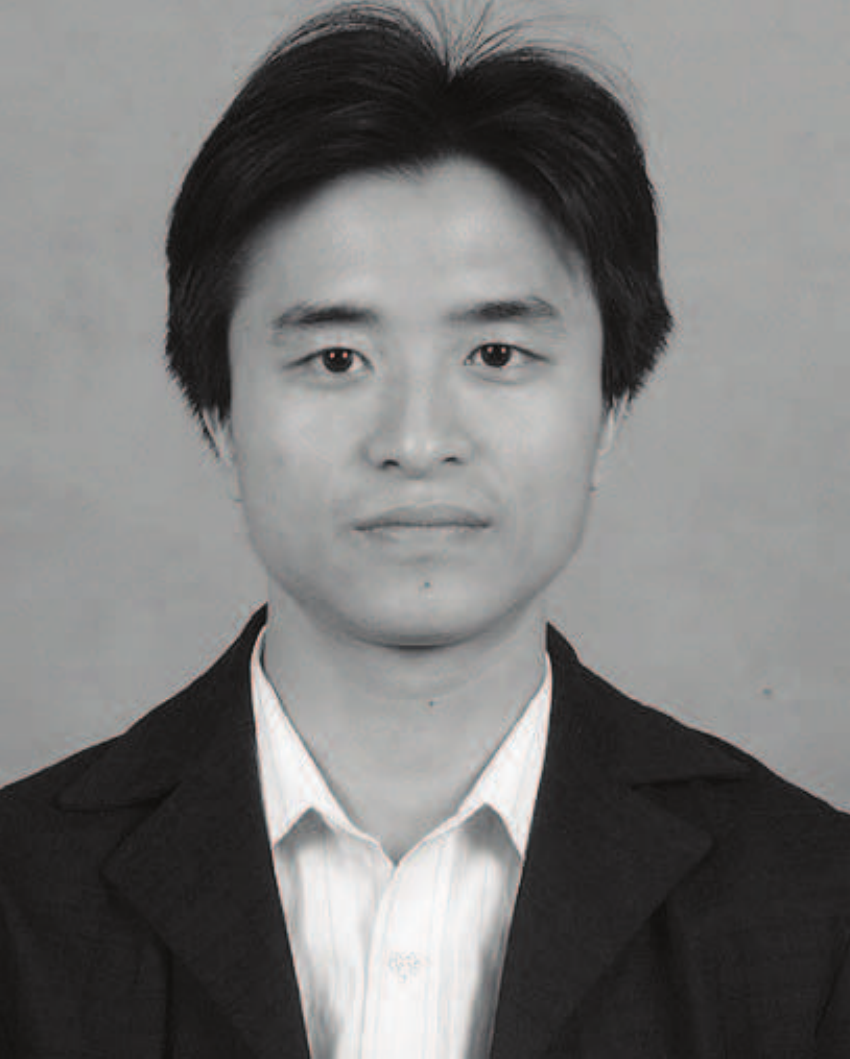}}]{Huaping Liu}
 Huaping Liu received the Ph.D. degree from Tsinghua University, Beijing, China, in 2004. \\He currently is an Associate Professor with the Department of Computer Science and Technology, Tsinghua University, Beijing, China. His research interests include robot perception and learning.\\ Dr. Liu serves as an Associate Editor of several journals including the IEEE ROBOTICS AND AUTOMATION LETTERS, Neurocomputing, Cognitive Computation, and some conferences including the International Conference on Robotics and Automation and the International Conference on Intelligent Robots and Systems. He also served as a Program Committee Member of RSS2016 and IJCAI2016.
 \end{IEEEbiography}

\begin{IEEEbiography}[{\includegraphics[width=1in,height=1.25in,clip,keepaspectratio]{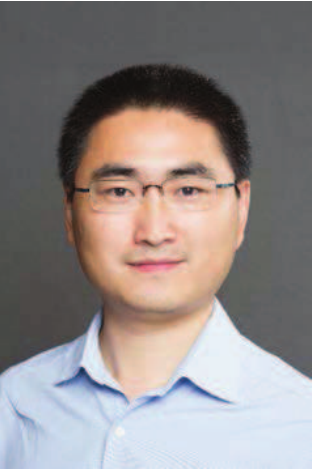}}]{Jun Liu}Jun Liu received the Ph.D. degree from University of Toronto, Toronto, Canada, in 2016. From 2017 to 2019, he was a Postdoc Fellow in the Dalio Institute of Cardiovascular Imaging, Cornell University, New York, USA. He currently is a teacher in Department of Mechanical Engineering, City University of Hong Kong, Hong Kong, China. His research interests include micro-nano robotics and medical image analysis and interaction.\\ His research has been recognized on the major robotics and automation conferences by winning multiple awards including the Best Student Paper Award and Best Medical Robotics Paper Finalist Award from the IEEE International Conference on Robotics and Automation in 2014, and the IEEE Transactions on Automation Science and Engineering Best New Application Paper Award in 2018.
\end{IEEEbiography}

\end{document}